%% file: iclr2022_conference.tex
\documentclass{article} 
\usepackage{iclr2022_conference,times}
\usepackage{graphicx}
\usepackage{multirow}
\usepackage{subcaption}
\usepackage[ruled,vlined]{algorithm2e}
\usepackage{wrapfig}
\SetKwInput{KwInput}{Input}  

\input{math_commands.tex}

\usepackage{hyperref}
\usepackage{url}

\title{MixNorm: Test-Time Adaptation through \\Online Normalization Estimation}

\author{Xuefeng Hu$^{1}$, Gokhan Uzunbas$^{2}$, Sirius Chen$^{2}$, Rui Wang$^{2}$, Ashish Shah$^{2}$,\\
\textbf{Ram Nevatia$^{1}$ \& Ser-Nam Lim$^{2}$}\\
$^{1}$ University of Southern California, $^{2}$ Facebook AI 
}

\iclrfinalcopy 
\begin{document}

\maketitle

\begin{abstract}
We present a simple and effective way to estimate the batch-norm statistics during test time, to fast adapt a source model to target test samples. Known as Test-Time Adaptation, most prior works studying this task follow two assumptions in their evaluation where (1) test samples come together as a large batch, and (2) all from a single test distribution. However, in practice, these two assumptions may not stand, the reasons for which we propose two new evaluation settings where batch sizes are arbitrary and multiple distributions are considered. Unlike the previous methods that require a large batch of single distribution during test time to calculate stable batch-norm statistics, our method avoid any dependency on large online batches and is able to estimate accurate batch-norm statistics with a single sample. The proposed method significantly outperforms the State-Of-The-Art in the newly proposed settings in Test-Time Adaptation Task, and also demonstrates improvements in various other settings such as Source-Free Unsupervised Domain Adaptation and Zero-Shot Classification.
\end{abstract}

\section{Introduction}\label{section-introduction}
    \input{sections/introduction_snl}
\section{Related Work}\label{section-related}
    \input{sections/related_snl}

\section{Test-Time Adaptation with Mix-Norm Layer}\label{section-method}
    \input{sections/method_snl}

\section{Experiments and Results}\label{section-experiment}
    \input{sections/experiments_snl}
\section{Conclusion}\label{section-discussion}
    We have developed a novel way to estimate the batch-norm statistics during test time, to fast adapt a source model to target test samples. Unlike the previous methods that require a large batch from a single distribution to calculate stable batch-norm statistics, our proposed method eliminates any dependency on online batches and is able to estimate accurate batch-norm statistics with single samples. The newly proposed method significantly outperformed the State-Of-The-Art models in the newly proposed real-world settings, and also demonstrated potential to improve zero-shot learning performance.
\newpage
\bibliography{references} 
\bibliographystyle{iclr2022_conference}

\newpage
\appendix
\section{Full Results on CIFAR-10C, ImageNet-C}
    \input{sections/appendix/full_result}

\section{Hyper Parameter Selection}
    \input{sections/appendix/hyper}
\section{Zero-Shot Classification Datasets}
    \input{sections/appendix/datasets}
\end{document}

%% file: math_commands.tex
\usepackage{amsmath,amsfonts,bm}

\def\eqref#1{equation~\ref{#1}}

\def\1{\bm{1}}

\DeclareMathAlphabet{\mathsfit}{\encodingdefault}{\sfdefault}{m}{sl}
\SetMathAlphabet{\mathsfit}{bold}{\encodingdefault}{\sfdefault}{bx}{n}



%% file: sections/introduction_snl.tex
\begin{wrapfigure}{ht}{0.5\textwidth}
  \begin{center}
    \includegraphics[width=0.5\textwidth]{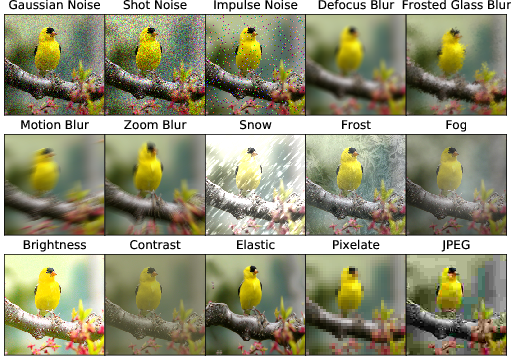}
  \end{center}
  \caption{Examples from ImageNet-C dataset \cite{hendrycks2019benchmarking}. ImageNet-C contains 75 corrupted copies of the original ImageNet-1K test set, with 5 severity levels and 15 different corruption types. This dataset is used to simulate the distribution shift between training domain (original) and test domains (corrupted).}
  \label{fig:imagenet}
\end{wrapfigure}

Deep learning models have brought tremendous performance improvements over many different domains (\cite{krizhevsky2012imagenet,he2016deep,deng2009imagenet}) in recent years. 
In spite of this, we often find that a well-trained model can perform unexpectedly poorly in real-world settings. Many factors play a role in this challenging problem; but one of the most probable causes stems from a domain shift from the training data (~\cite{quinonero2009dataset}). To study the domain shift in production environment and to evaluate the generalization ability of deep learning models, many benchmarks have been proposed (\cite{saenko2010adapting,peng2017visda,hendrycks2019benchmarking}). 
Figure \ref{fig:imagenet} demonstrates the samples from one of the benchmarks, ImageNet-C (\cite{hendrycks2019benchmarking}), in which synthesized corruptions are added to ImageNet (\cite{deng2009imagenet}) test images in an attempt to simulate domain shift. 

To increase deep learning models' robustness when there is domain shift, various approaches have been developed in different settings, such as Domain Adaptation (\cite{patel2015visual}), Unsupervised Domain Adaptation(\cite{wilson2020survey}), Source-Free Domain Adaptation(\cite{liang2020we}) and Fully Test-Time Adaptation (\cite{wang2020tent}). Among all those different settings, the task of Fully Test-Time Adaptation is one of the most challenging settings. With only access to a model that is pre-trained offline, the goal of Test-Time Adaptation (TTA) is to rapidly adapt the model to the test samples while making predictions. The task is essentially Domain Adaptation plus three more challenging conditions:
\begin{itemize}
    \item \textbf{Online}: The model gets updated during inference, and has access to each test sample only once.
    \item \textbf{Source-Free}: The model does not have access to training data after the offline training is done. It will only have access to the pre-trained weights. 
    \item \textbf{Unsupervised}: There is no label for the test data. 
\end{itemize}

Several works have studied the Test-Time Adaptation problem in recent years (\cite{li2016revisiting,sun2020test,wang2020tent}). Existing methods focus on adapting the Batch Normalization Layers (\cite{ioffe2015batch}) in the deep learning models, and have significantly improved deep learning model's performance against test-time domain shift. 

Such Batch-Norm Layer based methods make two underlying assumptions to estimate stable normalization statistics during evaluation: 1) test data come together in large batches;
2) all the test samples are from the same shifted distribution (single corruption types, single domain, etc). However, such assumptions might not be practical in the real world as the samples might come from an arbitrary distribution such as: varying sample sizes, different ways of collection, or different types of corruptions. Further, during inference, the AI system might not have the freedom to postpone the prediction in order to collect enough data to apply test-time adaptation algorithm on the incoming test samples. 

To eliminate the dependency on large batch size and sole distribution, we propose to replace the Batch-Norm Layers with a novel MixNorm operation that performs adaptive prediction of the normalization statistics from a single input. The MixNorm Layer considers not only global statistics calculated from all historical samples, but also local statistics calculated from incoming test sample and its spatial augmentations. Figure \ref{fig:all} illustrates our overall approach. The novel MixNorm method demonstrate significant improvement over the State-of-the-art methods TENT (\cite{wang2020tent}), especially in the two newly proposed evaluation protocols, where the all Test-Time Adaptation methods are evaluated at various batch sizes, and against mixture of test samples from different shifted domains. In addition to the Test-Time Adaptation task, the proposed MixNorm Layer demonstrates improvement over two other tasks, source-free Domain Adaptation and Zero-Shot Image Classification. 

Our contributions are:
\begin{itemize}
    \item We propose two new evaluation conditions for the task of Test-Time Adaptation, which avoid the impractical assumptions made by previous protocols. 
    \item We propose a novel and simple way to adaptively estimate the test-time normalization statistics from single sample.
    \item We performed extensive experiments that demonstrate the effectiveness of the new MixNorm Layer in various tasks, such as Test-Time Adaptation, Source-Free Unsupervised Domain Adaptation and Zero-Shot Image Classification. 
\end{itemize}

%% file: sections/related_snl.tex
Distribution shift between training domain and evaluation domain often causes a drop in the performance of deep learning models. In recent years, many proposals have been made to increase deep learning model's robustness against distribution shifts, a field of research known as Domain Adaptation. Domain Adaptation methods can be classified into different settings depending on factors like access to source (training) samples, existence of new or missing categories, availability of supporting samples in evaluation domain, etc. In this paper, we focus on settings where there is no access to the source (training) samples. In particular, we compare our new method with existing methods on Test-Time Adaptation, Source-Free Unsupervised Domain Adaptation and Zero-Shot Classification settings. 

\begin{figure}
    \centering
    \includegraphics[width=0.9\textwidth]{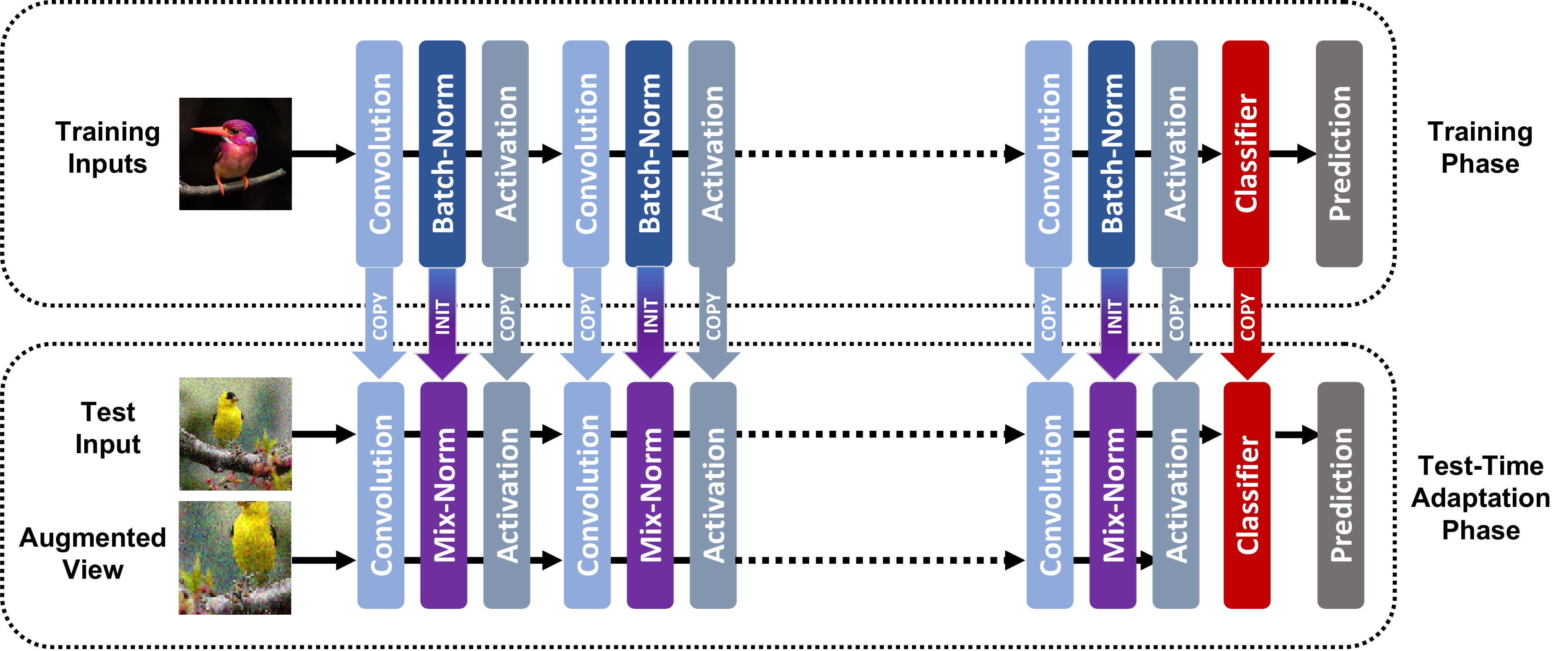}
    \caption{Method overview for Test-Time Adaptation with MixNorm layers. Before the inference, we replace all the Batch-Norm layers in the pre-trained model with the new proposed MixNorm layers. The MixNorm layers are initialized by the training set (source) statistics (means and variances) and the corresponding weights and biases from the pre-trained Batch-Norm layers. During inference, each input is paired with another augmented view for calculation in the MixNorm layers and the final prediction is made only on the original input.}
    \label{fig:all}
\end{figure}

\subsection{Test-Time Adaptation}
The task of Test-Time Adaptation (TTA) is one of the most challenging settings in the field of Domain Adaptation, as it requires rapid adaptation while making predictions, without access to anything but the pre-trained model and testing samples. BN (\cite{li2016revisiting}) was one of the earliest work that explores the TTA task. It focuses on adapting the Batch-Norm Layers in deep learning models. A traditional Batch-Norm module collects running means and variances at training time from the training data, and uses the collected means and variance at test time. BN argued that when there are distribution shifts between training and testing data, it should not keep using the same training statistics. Instead, the model should collect statistic directly from the online batch. BN demonstrates that this simple approach of using online statistics is effective in handling the distribution shifts caused by corruptions. However, one clear drawback of the method is that it requires relative large batches to obtain stable online statistics. Following the work of BN (\cite{li2016revisiting}), TTT (\cite{sun2020test}) proposed to update not only the batch-norm statistics, but also the affine weights and biases, which are learnable weights in Batch-Norm Layers that further re-scale the normalized output. TTT performs a single step gradient update over the batch-norm parameters after each prediction, with an unsupervised loss defined by rotation prediction task. TENT (\cite{wang2020tent}) further proposed to use entropy loss instead of rotation prediction loss and achieved state-of-the-art performance on CIFAR-10C, CIFAR-100C and ImageNet-C datasets.

Most of the existing works in TTA follow the same evaluation protocol proposed by \cite{li2016revisiting}, where the models are adapted using large batches of test samples coming from a single type of corruption. However, such an assumption is not realistic in practice, because the test samples in the real world could come in any batch size (in particular, real time high performance inference platforms often process one input at a time, \cite{Anderson2021FirstGenerationIA}), or from any shifted distribution. 

\subsection{Source-Free Unsupervised Domain Adaptation}
Similar to the task of TTA, in Source-Free Unsupervised Domain Adaptation (UDA), the models only have access to pre-trained models and test samples. Different than the TTA setting, Source-Free UDA models have access to the evaluation samples as many times as needed during adaptation.

SHOT \cite{liang2020we} is currently the state of the art in Source-Free UDA, and also one of the first to propose UDA without needing any source data. In SHOT, the parameters are fine-tuned with a self-supervised loss that minimize the prediction entropy in the target samples, while keeping the classification head intact. SHOT demonstrates that it is possible to get reasonable performance on standard benchmarks such as Office (\cite{saenko2010adapting}), Office-Home (\cite{venkateswara2017deep}) and VisDA (\cite{peng2017visda}). 

Another method called TENT (\cite{wang2020tent}) also has shown potential for the Source-Free UDA task with promising performance on digit datasets (e.g. adapting SVHN model to MNIST/MNIST-M/USPS). TENT cannot reach SHOT's performance when scaled to much larger scale datasets such as VisDA (\cite{peng2017visda}), but TENT adapts to target domain inputs with much cheaper operations and sees each test sample only once during testing.

\begin{minipage}[t]{0.48\textwidth}
\begin{algorithm}[H]
\SetAlgoLined
\KwInput{
\\Input Feature:  $F\in R^{D\times H\times W}$
\\Augmentation Features of the same sample: $F'\in R^{N\times D\times H\times W}$
\\Training Set Mean $\mu^0$ and Variance $\sigma^0$
\\Exponential Moving Speed $\tau=0.001$
\\Mixing Scale $m=0.05$
\\Weights, Bias: $\alpha,\beta$
}
\textbf{Procedure:}\\
// Update $\mu$ and $\sigma$ with most recent sample \\
$\mu^{t+1} \gets (1-\tau)\mu^t + \tau F.mean(1,2)$ \\
$\sigma^{t+1} \gets (1-\tau)\sigma^t + \tau (F-\mu).var(1,2))$\\
// Get global statistics\\
$\mu_{global} \gets \mu^t $\\
$\sigma_{global} \gets \sigma^t $\\
// Get local statistics from Augmentations\\
$\mu_{local} \gets F'.mean(0,2,3)$\\
$\sigma_{local} \gets (F'-\mu_{local}).var(0,2,3)$\\
// Mix Global and Local\\
$\mu_{mixed} \gets (1-m)\mu_{global} + m\mu_{local}$\\
$\sigma_{mixed} \gets (1-m)\sigma_{global} + m\sigma_{local}$ \\
// Calculate shifted f\\
$F \gets \alpha \frac{F-\mu_{mixed}}{\sigma_{mixed}}+\beta$\\
return $F$\\
\caption{Mix-Norm Update}
\label{algorithm1}
\end{algorithm}
\end{minipage}
\hfill
\begin{minipage}[t]{0.48\textwidth}
\begin{algorithm}[H]
\SetAlgoLined
\KwInput{
\\Batch of Features: $F\in R^{B\times D\times H\times W}$
\\Augmentated Features of the same batch: $F'\in R^{B\times N\times D\times H\times W}$
\\Training Set Mean $\mu^0$ and Variance $\sigma^0$
\\Max Moving Speed $\tau_{max} = 0.9$
\\Mixing Scale $m=0.05$
\\Weights, Bias: $\alpha,\beta$
}
\textbf{Procedure:}\\
// Adjust $\tau$ based on Batch Size $B$\\
$\tau \gets \tau_{max}10^{-\frac{3}{B}}$\\
// Update $\mu$ and $\sigma$ with most recent sample \\
$\mu^{t+1} \gets (1-\tau)\mu^t + \tau F.mean(0,1,2)$ \\
$\sigma^{t+1} \gets (1-\tau)\sigma^t + \tau (F-\mu).var(0,1,2)$\\
// Update global statistics with current batch\\
$\mu_{global} \gets \mu^t $\\
$\sigma_{global} \gets \sigma^t $\\
// Get local statistics from Augmentations\\
$\mu_{local} \gets F'.mean(0,1,3,4)$\\
$\sigma_{local} \gets (F'-\mu_{local}).var(0,1,3,4)$\\
// Mix Global and Local\\
$\mu_{mixed} \gets (1-m)\mu_{global} + m\mu_{local}$\\
$\sigma_{mixed} \gets (1-m)\sigma_{global} + m\sigma_{local}$ \\
// Calculate shifted f\\
$F \gets \alpha \frac{F-\mu_{mixed}}{\sigma_{mixed}}+\beta$\\
return $F$\\
\caption{MixNormBN: Combines MixNorm and BatchNorm}
\label{algorithm2}
\end{algorithm}
\end{minipage}

\subsection{Zero-Shot Classification}
Unlike Test-Time Adaptation or Source-Free UDA, where methods are developed to close the domain gap during evaluation stage, the goal of Zero-Shot Classification (\cite{xian2018zero}) is to increase the generalization ability during the training stage so that it can be applied to any test domain without adaptation. Without access to any support samples from the new categories, Zero-Shot Classification methods typically take advantage of cues from syntax information such as label embedding (\cite{frome2013devise}) or attributes (\cite{lampert2013attribute}) to establish knowledge about the new categories. 

Recently, along with the success in large-scale pre-training (\cite{brown2020language}), self-supervised learning  and contrastive learning (\cite{chen2020simple}), CLIP (\cite{radford2021learning}) has brought a huge improvement over almost all of the common benchmark datasets. CLIP collects 400 millions image-caption pairs and learns a joint representation space of visual and language representations. With the joint representation space, CLIP has the ability to do Zero-Shot Classification by directly looking at the potential category names, and its Zero-Shot Classification performance is close to that of fully supervised models on many datasets including ImageNet (\cite{deng2009imagenet}).

%% file: sections/method_snl.tex
In this section, we define our new Mix-Norm layer and provide details on how it can be utilized for closing domain gap during test time on Deep Neural Networks (DNN). Given a pre-trained DNN, we replace its existing Batch-Norm layers (\cite{ioffe2015batch}). During test time, for arbitrary sized batches our new MixNorm layer calculates empirical normalization statistics from two sources:

\begin{itemize}
    \item Global Statistics: The Batch-Norm layers from the pre-trained model store the statistics from the training distribution. Our MixNorm layer uses this training statistics as a global anchor point. The global statistics, initialized by the training statistics, is updated by an exponential moving average of the online test sample. 
    \item Local Statistics: We create additional augmented views of the test sample to form a small augmentation batch. We calculate the empirical statistics from this augmentation batch to capture the local distribution shift of the test sample. The augmentations are spatial, including random-resized-crop and random-flipping. In practice, we only use one additional augmentation as it has the best performance.  
\end{itemize}

In a typical deep learning model such as ResNet (\cite{he2016deep}) or WideResNet (\cite{zagoruyko2016wide}), there will be multiple batch-norm layers across the whole backbone network. Figure \ref{fig:mixnorm} below, illustrates our proposed operation in only one such layer. As shown earlier in Figure \ref{fig:all}, each sample $X$ comes along with its augmented view $X'$. After the encoding layers and right before our MixNorm layer, input images $X, X'\in \mathbb{R}^{3\times224\times224}$ become feature tensor $F,F'\in\mathbb{R}^{D\times H\times W}$. For simplicity, we denote the $i$-th row, $j$-th column of $F, F'$ to be $F_{i,j},F'_{i,j}\in\mathbb{R}^{D}$.

As depicted in Figure \ref{fig:mixnorm}, our global statistics $\mu^t, \sigma^t$ are initialized with train-time statistics, copied from the batch-norm layers in the pre-trained network:
\begin{align*}
    &\mu^0 = \mu^{training} & \sigma^0 = \sigma^{training},
\end{align*}
and they will be slowly updated by the statistics on each new sample. At $(t+1)$-th test example:
\begin{align*}
    &\mu^{t+1} = (1-\tau)\mu^{t}+\tau \mu &\sigma^{t+1} = (1-\tau)\sigma^{t}+\tau \sigma,
\end{align*}
where $\tau$ is the exponential moving speed, and $\mu$ and $\sigma$ are the means and variances from $F$:
\begin{align*}
    & \mu = \frac{\sum_{i=0}^H\sum_{j=0}^W F_{ij}}{HW} &\sigma = \frac{\sum_{i=0}^H\sum_{j=0}^W (F_{ij}-\mu)^2}{HW}.
\end{align*}
On the other hand, we calculate the local statistics $\mu_{local},\sigma_{local}$ by examining both $F,F'$:
\begin{align*}
    & \mu_{local} = \frac{\sum_{F^*\in\{F,F'\}}\sum_{i=0}^H\sum_{j=0}^W F^*_{ij}}{2HW} & \sigma_{local}= \frac{\sum_{F^*\in\{F,F'\}}\sum_{i=0}^H\sum_{j=0}^W (F^*_{ij}-\mu_{local})^2}{2HW}.
\end{align*}
Renaming $\mu^{t+1},\sigma^{t+1}$ to $\mu_{global},\sigma_{global}$, we obtain our final statistics:
\begin{align*}
    & \mu_{mixed}= (1-m)\mu_{global} + m\mu_{local} & \sigma_{mixed}= (1-m)\sigma_{global} + m\sigma_{local}.
\end{align*}
Finally, with $\mu_{mixed}, \sigma_{mixed}$, we can perform the normalization operation, and forward the resulting $F,F'$ to next layers:
\begin{align*}
    & F= \alpha\frac{F-\mu_{mixed}}{\epsilon+\sqrt{\sigma_{mixed}}}+\beta & F'= \alpha\frac{F'-\mu_{mixed}}{\epsilon+\sqrt{\sigma_{mixed}}}+\beta,
\end{align*}
where $\alpha,\beta,\epsilon$ are the parameters copied from the pre-trained network batch-norm layers.

\begin{figure}[!ht]
    \centering
    \includegraphics[width=0.9\textwidth]{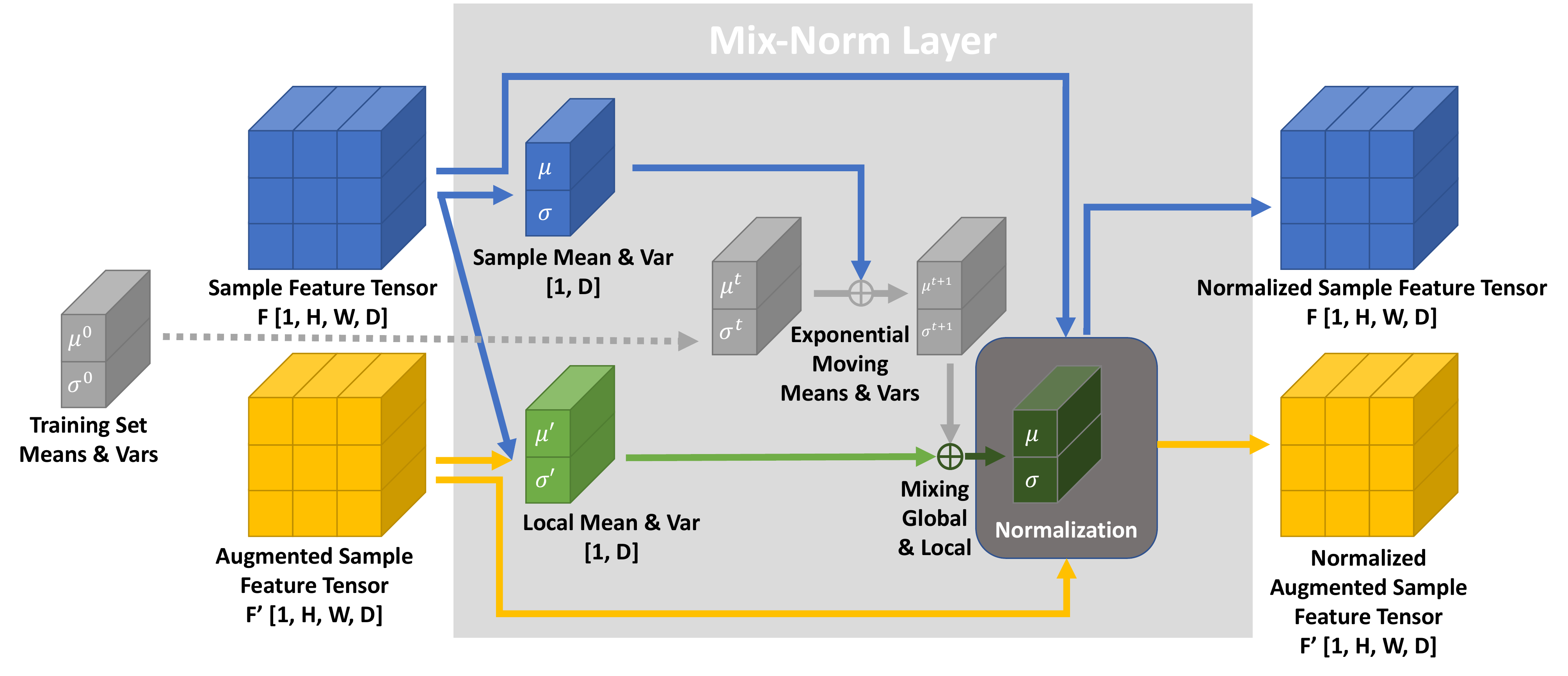}
    \caption{Illustration of the MixNorm Layer. Inside the MixNorm layer we estimate the normalization statistics by looking at both global statistics from historical data and local statistics from augmented views.}
    \label{fig:mixnorm}
\end{figure}

While the above process describes the procedure in Algorithm \ref{algorithm1}, we also present an variant of MixNorm in Algorithm \ref{algorithm2}. Different from the vanilla MixNorm which only consider single sample, Algorithm \ref{algorithm2} updates the global statistics with batch statistics, and have an adaptive moving speed $\tau$ which approximates Batch-Norm behavior for large batch size, and MixNorm behavior for small batch size.

%% file: sections/experiments_snl.tex

\textbf{Datasets}. For the Test-Time Adaptation task, we follow TENT (\cite{wang2020tent}) and evaluate our method over the standard benchmarks CIFAR-10C and ImageNet-C (\cite{hendrycks2019benchmarking}). CIFAR-10C includes 75 subsets that are composed of 15 types of corruptions at 5 severity levels. It comprises the 10,000 images, each of size 32x32 pixels, from the 10 classes in the CIFAR10 test set, but with each modified by the corresponding corruption type at certain severity level. Similarly, ImageNet-C is composed of a set of 75 common visual corruptions, applied to the original ImageNet validation set, with 50,000 images of 224x224 pixels from 1000 classes. 

For Source-Free Unsupervised Domain Adaptation task, we follow the experiment setup from \cite{wang2020tent}, and use SVHN (\cite{svhn}) as the source domain and MNIST (\cite{mnist})/MNIST-M (\cite{mnistm})/USPS (\cite{svhn}) as the target domains. SVHN, MNIST, MNIST-M and USPS are all 10-class digits datasets, with training/testing data sizes of 73257/26032, 60000/10000, 60000/10000 and 7291/2007 respectively. 

For Zero-Shot Image Classification, we follow the setup from CLIP (\cite{radford2021learning}) and evaluate our method on CIFAR-10/100 (\cite{krizhevsky2009learning}), STL-10 (\cite{stl10}), Stanford Cars (\cite{cars}) and Food101 (\cite{food}). Details of these datasets are provided in the Appendix C. 

\textbf{Implementation Details}. For Test-Time Adaptation experiments, we follow the optimization setting from TENT (\cite{wang2020tent}), which uses Adam optimizer on CIFAR-10C with learning rate 0.001 and SGD optimizer on ImageNet-C with learning rate 0.00025, both optimized by the entropy minimization loss on output logits as described in \cite{wang2020tent}. For CIFAR-10C we compare with TENT on both Wide-ResNet-28 and Wide-ResNet-40 (\cite{zagoruyko2016wide}) and for ImageNet-C we compare with TENT on ResNet-50 (\cite{he2016deep}) following the official public model provided by the TENT repository\footnote{\url{https://github.com/DequanWang/tent}}. 

For Source-Free Unsupervised Domain Adaptation experiments, we replicated TENT (\cite{wang2020tent}) and BN (\cite{li2016revisiting}) methods using the pre-trained SVNH model repository\footnote{\url{https://github.com/aaron-xichen/pytorch-playground}} since the original ResNet-26 model reported in TENT (\cite{wang2020tent}) was not public. For hyper-parameter selection, on CIFAR-10C/ImageNet-C we use the ``Gaussian Noise at Severity 5" set to choose the hyper parameters, and then use the same hyper-parameters during all experiments. For Source-Free UDA and Zero-Shot Classification experiments, as there is no trivial way to split a validation set, we just report the optimal hyper-parameters for each dataset. The details of the selected hyper-parameters will be provided in Appendix B. For all the experiments, we use RandomResizedCrop and RandomFlip as the augmentation methods.

\subsection{Test-Time Adaptation}
To understand the merit of our new method in real world scenarios, we introduce two new conditions during evaluations. In TENT, error rates for each of the 15 corruption types at severity 5 are reported with a fixed batch size of 200 on CIFAR-10C and 64 on ImageNet-C. In addition to this standard protocol, we compare our method with TENT under two new conditions: 

\textbf{1) Different Batch Sizes}. In additional to the fixed 200/64 batch size in the previous protocol, we also compare our method with TENT at batch sizes of 1, 5, 8, 16, 32, 64, 100, 200; 

\textbf{2) Mixed Distribution}. In additional to the standard protocol where the model's final error rate is averaged from 15 separate evaluations, each of which contains only one type of corruption, we further evaluate the models with all 15 corruptions types shuffled and mixed together.  

\begin{figure}[ht]
\begin{subfigure}{.5\textwidth}
  \centering
  \includegraphics[width=\linewidth]{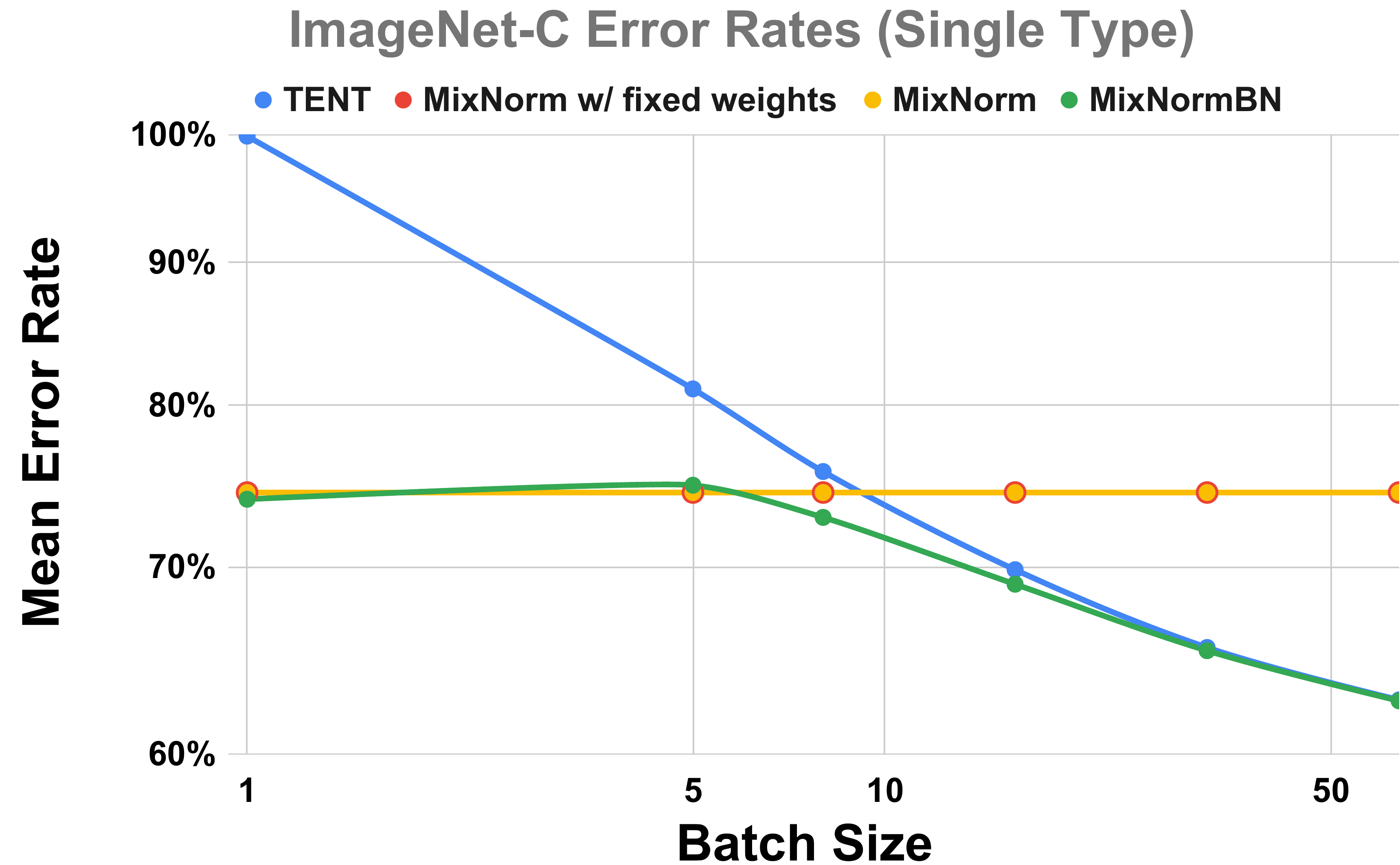} 
  \label{fig:sub-first-1}
\end{subfigure}
\begin{subfigure}{.5\textwidth}
      \centering
      \includegraphics[width=\linewidth]{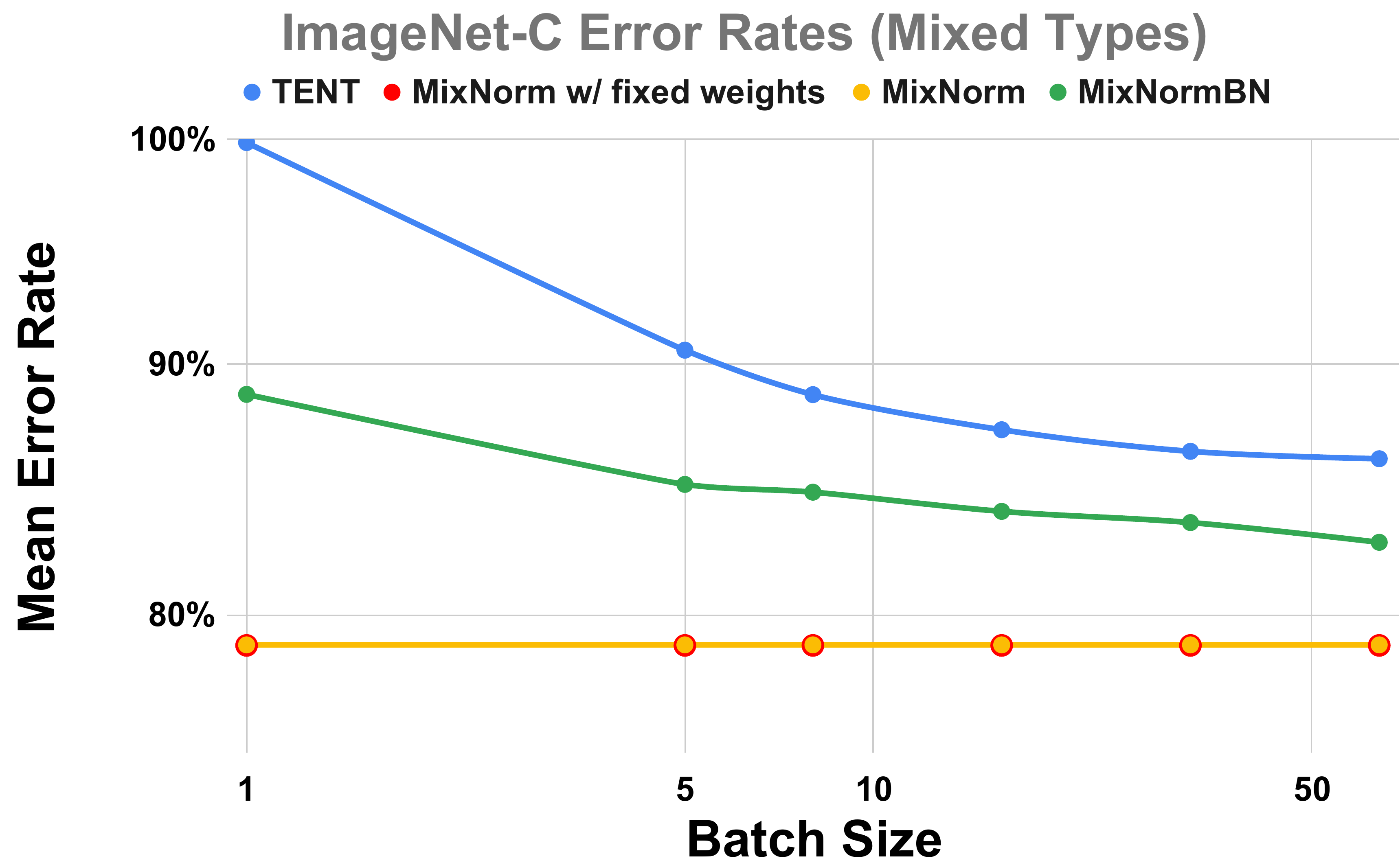}
      \label{fig:sub-second-1}
    \end{subfigure}
    \begin{subfigure}{.5\textwidth}
      \centering
      \includegraphics[width=\linewidth]{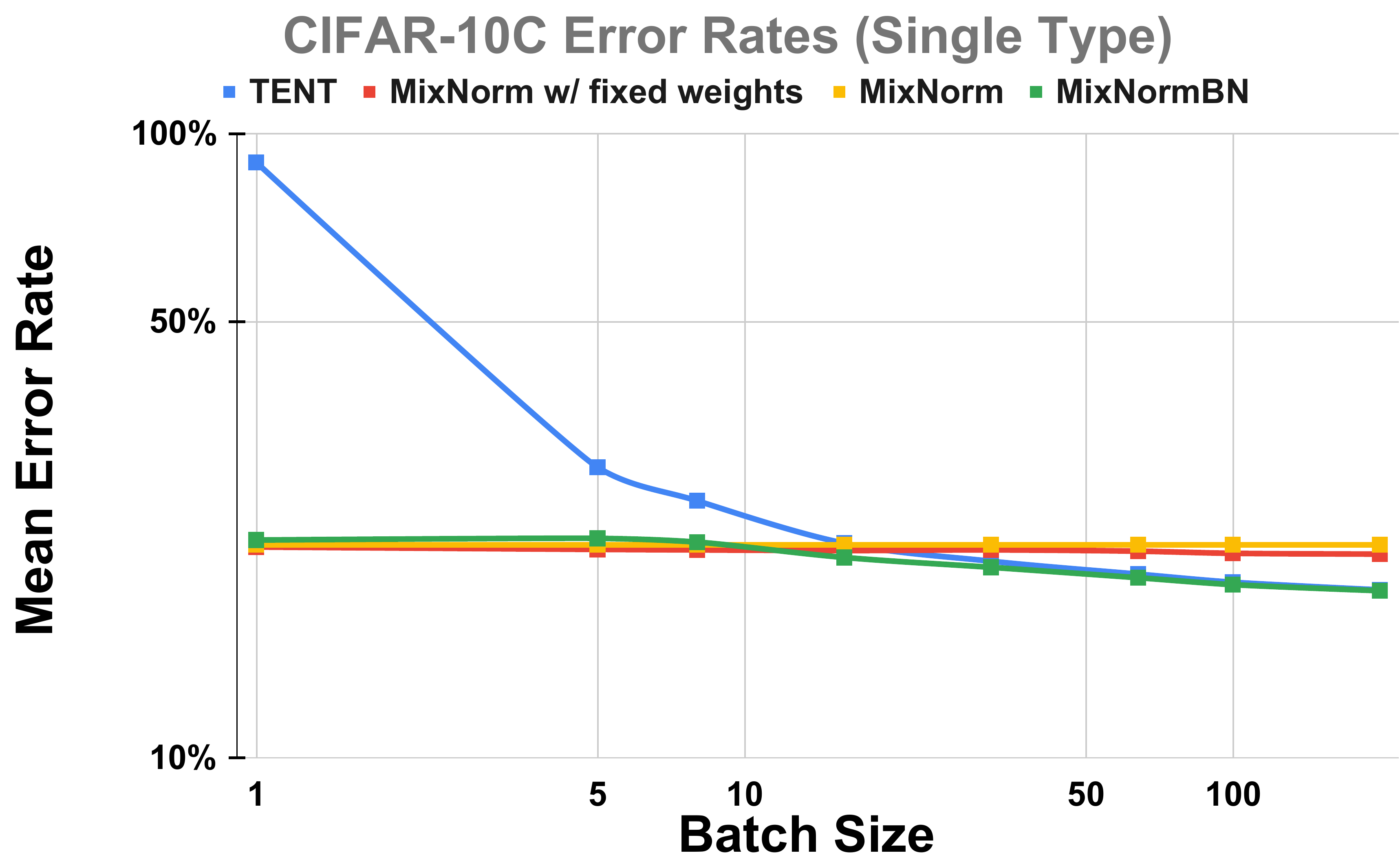}
      \label{fig:sub-first-2}
    \end{subfigure}
    \begin{subfigure}{.5\textwidth}
      \centering
      \includegraphics[width=\linewidth]{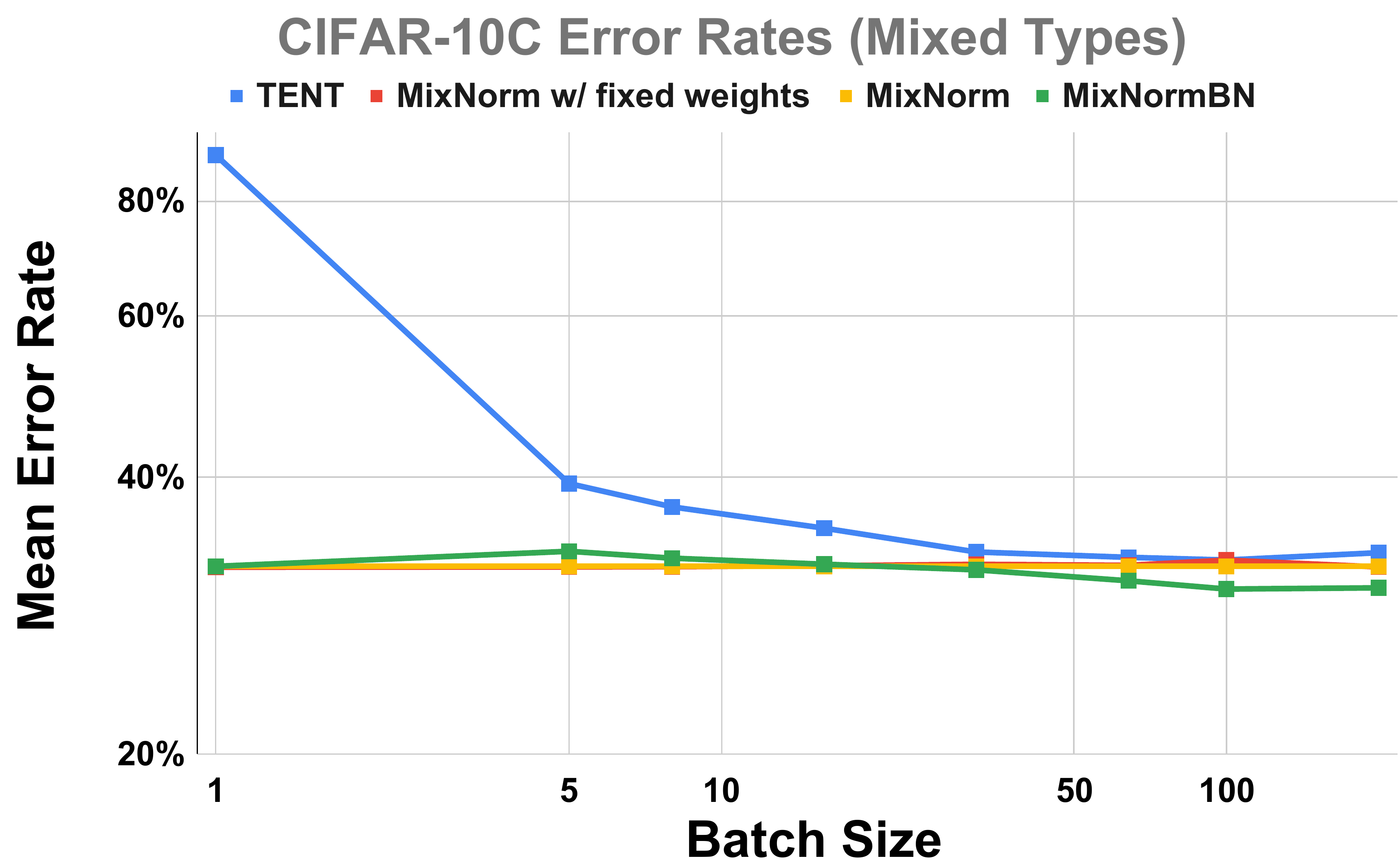}
      \label{fig:sub-second-2}
    \end{subfigure}
    \caption{Comparison of MixNorm, with and without learnable affine weights (Algorithm \ref{algorithm1}), and MixNormBN (Algorithm \ref{algorithm2}) methods to the baseline method TENT in two different test time settings i) test samples come only from a single type of corruption; ii) test samples come from mixed types of corruptions. For the Single Type experiments, we report average error rates over all 15 corruption types at severity 5.  Numerical results are reported in the Appendix A.} 
    \label{fig:robustness}
    \end{figure}

Figure \ref{fig:robustness} shows our main results. 
Blue curves represent the SOTA model TENT, the red and orange curves represent MixNorm model with and without learnable affine weights $\alpha$ and $\beta$. Green curves represent the performance of MixNormBN as described in Algorithm \ref{algorithm2}. Tent's performance drops as the batch size decreases, while MixNorm's performance stays stable for any batch size. Additional to the improvement to small batch size, Figure \ref{fig:robustness} also shows the advantages of MixNorm when test samples come from different distributions. 
In TENT's testing protocol, all test samples are collected from the same corruption distribution, which makes it easier to obtain reliable normalization statistics especially when the batch size is large. However, when tested with batches of mixed inputs sampled from different corruption types, such assumption does not hold and Tent's performance drops significantly compared to MixNorm. This result holds even when tested with larger batch size (up to 64 on ImageNet-C and up to 200 on CIFAR-10C). We also observe that when batch size is large enough and all samples come from a similar distribution, TENT might have an advantage over MixNorm; because TENT can get to estimate stable normalization statistics from empirical samples. In order to take advantage of large batch size, when we test our MixNormBN method (green curves), again we report consistently better scores than TENT at all batch sizes. 

\begin{figure}[ht]
\begin{subfigure}{.5\textwidth}
  \centering
  \includegraphics[width=\linewidth]{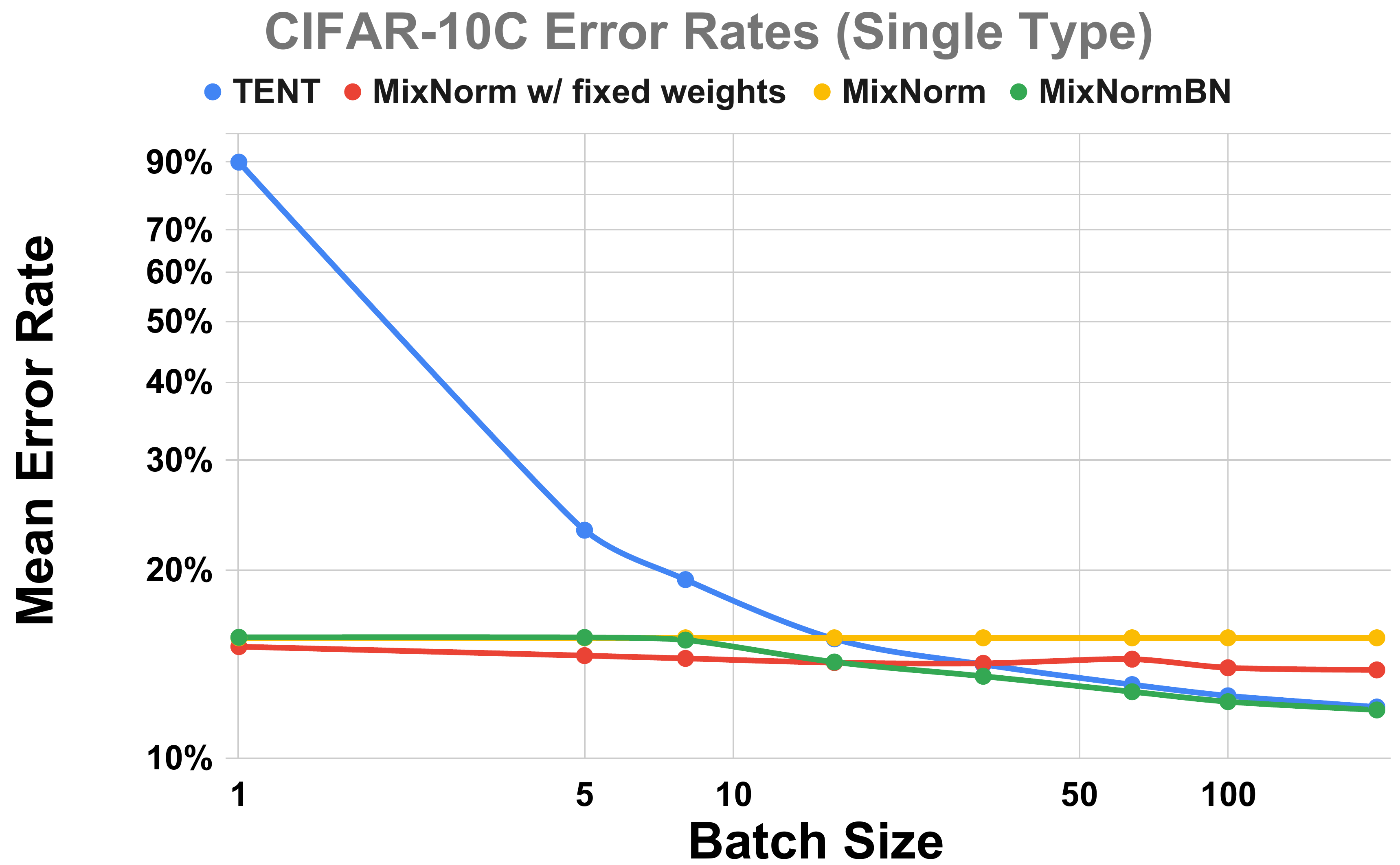} 
\end{subfigure}
\begin{subfigure}{.5\textwidth}
  \centering
  \includegraphics[width=\linewidth]{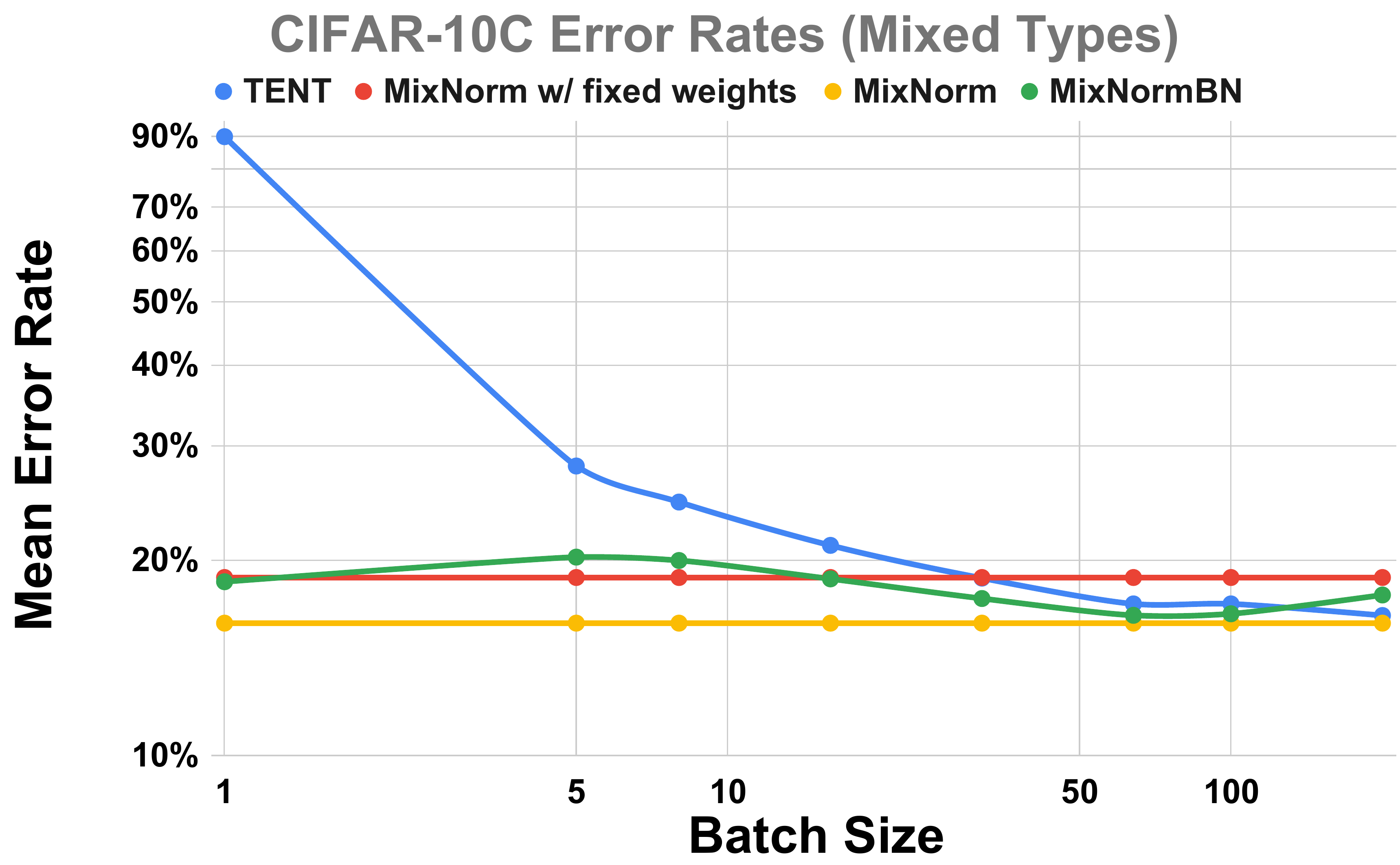}  
\end{subfigure}
\caption{Comparison of our MixNorm and MixNormBN methods to the baseline method TENT, with a different backbone architecture (WideResNet40). Numerical results are reported in the Appendix A.}
\label{fig:arch}
\end{figure}
            
\subsection{Ablation Studies}
In this section, we present the results of several ablations on architecture selection, ways of collecting normalization statistics and number of augmentations.
    
\textbf{Architectures:}
Figure \ref{fig:arch} provides the results on CIFAR-10C where both TENT and MixNorm adopt a more powerful backbone, WideResNet40, as opposed to the WideResNet28 used in Figure \ref{fig:robustness}. Along with the ResNet-50 used in the ImageNet-C experiments, we demonstrate the effectiveness of MixNorm on architecture with different depth and width. 
    
\textbf{Normalization Statistics Estimation:}
In Table \ref{fig:ablation}, we present an ablation study on different ways to collect normalization parameters. They include: 1) \textit{Instance Norm}: each feature is normalized by itself along the spatial dimensions; 2) \textit{Augmentation (Local) Norm}: features are normalized within the local batch of original samples and corresponding augmented views, i.e., only local statistics are used; 3) \textit{Fixed Global Norm}: each feature is normalized by the fixed statistics stored in the pre-trained weights learned during training; 4) \textit{Moving Global Norm}: each feature is normalized by the exponential moving statistics, i.e., only global statistics are used.

As shown in the table row 1 and 2, Augmentation Norm improves on Instance Norm by having augmented samples in addition to the original samples, but both of them perform poorly due to the absence of global statistics. Rows 3 and 4 show that Moving Global Norm works better than Fixed Global Norm, which is expected due to the existence of domain shift. Rows 5, 6, 7 indicate that combining the global and local statistics boosted performance by a significant scale. Row 8 indicates that learnable affine weights, $\alpha$ and $\beta$, do not seem to provide additional boost to the performance observed in TENT, possibly because MixNorm can already make accurate prediction so it does not need to be adjusted by the affine weights. 

\textbf{Augmentations:}
Table \ref{fig:augmentation} shows that one augmentation is enough to get good performance, and the number of augmentations in local statistics calculation does not have significant influence on performance. Therefore, we just use one augmentation for faster computation. 

\begin{table}[!ht]
\begin{tabular}{ccccccllll}
\cline{1-6}
\multirow{2}{*}{No} & \multirow{2}{*}{Experiment Name} & \multicolumn{3}{c}{Way of Collecting Means \& Vars}                & \multirow{2}{*}{Error Rate} &\\ \cline{3-5}
                       &          & Moving Speed         & Global Scale         & Local Scale          &                             &\\ \cline{1-6}
1& Instance Norm                    & 0                    & 0                    & 1                    & 78.21\%                     &\\
2& Augmentation (Local) Norm        & 0                    & 0                    & 1                    & 68.73\%                     &\\ \cline{1-6}
3& Fixed Global Norm                & 0                    & 1                    & 0                    & 50.50\%                     &\\
4& Moving Global Norm               & 0.001                & 1                    & 0                    & 38.45\%                     &\\ \cline{1-6}
5& Mixed Norm                       & 0.001                & 0.95                 & 0.95                 & 34.45\%                     &\\
6& Mixed Norm                       & 0.001                & 0.9                  & 0.1                  & 32.66\%                     &\\
7& Mixed Norm                       & 0.001                & 0.8                  & 0.2                  & \textbf{31.96\%}            &\\ \cline{1-6}
8& Mixed Norm  w/ fixed params                    & 0.001                & 0.8                  & 0.2                  & \textbf{32.01\%}            &\\
\multicolumn{1}{l}{} & \multicolumn{1}{l}{}             & \multicolumn{1}{l}{} & \multicolumn{1}{l}{} & \multicolumn{1}{l}{} & \multicolumn{1}{l}{}        &
\end{tabular}
 \caption{Ablation studies on different ways to collect normalization statistics, tested with CIFAR-10C, with a mixed set of corruptions.}
    \label{fig:ablation}
\end{table}

\begin{table}[!ht]
\begin{tabular}{c|cccccc} 
Number of Augmentations & \textbf{1}       & 3       & 5       & 7       & 9       & 15      \\ \hline
ImageNet-C Error Rate             & \textbf{58.16\%} & 58.33\% & 58.35\% & 58.44\% & 58.49\% & 58.56\%
\end{tabular}
\caption{Ablation studies on the effect of number of augmentations to MixNormBN method when collecting normalization statistics, tested with ImageNet-C and ResNet-50, with a mixed set of corruptions.}
    \label{fig:augmentation}
\end{table}

\subsection{Test-Time Adaptation in Zero-Shot Learning Setting}
For Zero-Shot Classification, samples are evaluated separately at batch size of 1. Therefore, we use the regular MixNorm with fixed affine weights for comparison. We replace the Batch-Norm layers from the pre-trained CLIP models, with backbone architectures of ResNet-50, and then use the updated ResNet-50 model to calculate the visual features in Zero-Shot Classification. The results in Table \ref{clip-rn} shows MixNorm improves the CLIP's zero-shot learning performance on CIFAR-100, Stanford-Cars and CIFAR-10 and is on par with CLIP on Food-101 and STL10. This experiment aims to show MixNorm can bring reasonable improvements even over extremely large-scale pre-training models. Interestingly, the results show that our new method can bring larger improvements when the CLIP has a lower performance, which implicitly could be indication of the presence of larger domain shifts. 

\begin{table}[!ht]
\centering
\begin{tabular}{c|ccccc}
Model       & CIFAR-100 & Stanford-Cars & CIFAR-10 & Food-101 & STL10   \\ \hline
CLIP (RN50) & 40.97\%   & 53.80\%       & 72.26\%  & \textbf{78.99\%}  & \textbf{94.84\%} \\
MixNorm  & \textbf{45.13\%}   & \textbf{54.01\%}  &  \textbf{75.36\%}  & 78.78\%  & 94.67\%
\end{tabular}
\caption{Zero-Shot Image Classification Accuracy. We replaced the Batch-Norm layers in the pre-trained CLIP ResNet-50 weights and tested it under the zero-shot setting on the standard benchmarks. Result shows MixNorm improved the CLIP performance}
    \label{clip-rn}
\end{table}
    
\subsection{Source-Free Unsupervised Domain Adaptation}
\begin{table}[!ht]
\begin{tabular}{c|ccc}
Model & SVHN $\rightarrow$ MNIST & SVHN$\rightarrow$USPS & SVHN$\rightarrow$MNIST-M \\ \hline
Source               & 23.84\%                  & 37.02\%                 & 45.45\%                    \\
BN (\cite{li2016revisiting})              & 22.70\%                  & 36.87\%                 & 44.04\%                    \\
TENT (\cite{wang2020tent})                 & 8.17\%                   & 34.63\%                 & 42.28\%                    \\
MixNormBN        & \textbf{7.95\%}          & \textbf{25.71\%}        & \textbf{42.19\%}          
\end{tabular}
\caption{Error rate for Source-Free Unsupervised Domain Adaptation experiment on Digits Sets from SVHN to MNIST/MNIST-M/USPS.}
\label{uda}
\end{table}
For Source-Free UDA experiments, as we compare to TENT (\cite{wang2020tent}) and BN (\cite{li2016revisiting}), both of which use large batch sizes (200), we use MixNormBN for a fair comparison. 
As shown in Table \ref{uda}, MixNormBN outperforms both TENT and BN by a significant margin on all three datasets, indicating the effectiveness of MixNormBN in the task of Source-Free UDA, for other types of domain shift that are not just corruptions. 

%% file: sections/appendix/full_result.tex
In Table \ref{1},\ref{2}, \ref{3},\ref{4} we present the exact error rates shown in Figure \ref{fig:robustness}. In Table \ref{5},\ref{6} we present the error rates shown in  Figure \ref{fig:arch}.
\begin{table}[ht]
\centering
\begin{tabular}{c|cccc}
Batch Size & TENT    & MixNorm w/ Fixed Parameters & MixNorm & MixNormBN \\ \hline
1          & 90.00\% & 21.76\%                     & 21.94\% & 22.33\%   \\
5          & 29.21\% & 21.57\%                     & 21.94\% & 22.47\%   \\
8          & 25.81\% & 21.53\%                     & 21.94\% & 22.16\%   \\
16         & 22.07\% & 21.48\%                     & 21.94\% & 20.94\%   \\
32         & 20.65\% & 21.53\%                     & 21.94\% & 20.19\%   \\
64         & 19.69\% & 21.43\%                     & 21.94\% & 19.43\%   \\
100        & 19.11\% & 21.26\%                     & 21.94\% & 18.94\%   \\
200        & 18.58\% & 21.20\%                     & 21.94\% & 18.52\%  
\end{tabular}
\caption{\textbf{Single Distribution} Error Rates of TENT, MixNorm (Algorithm \ref{algorithm1}) with and without fixed affine parameters, MixNormBN (Algorithm \ref{algorithm2}) on \textbf{CIFAR-10C} datasets. All scores are average error rates from all 15 corruption datasets at severity level 5. All methods adopt \textbf{Wide-ResNet-28} as backbone architectures.}
\label{3}
\end{table}

\begin{table}[ht]
\centering
\begin{tabular}{c|cccc}
Batch Size & TENT    & MixNorm w/ Fixed Parameters & MixNorm & MixNormBN \\ \hline
1          & 89.69\% & 31.94\%                     & 32.01\% & 32.01\%   \\
5          & 39.39\% & 31.96\%                     & 32.01\% & 33.24\%   \\
8          & 37.15\% & 31.98\%                     & 32.01\% & 32.67\%   \\
16         & 35.22\% & 32.06\%                     & 32.01\% & 32.18\%   \\
32         & 33.18\% & 32.17\%                     & 32.01\% & 31.73\%   \\
64         & 32.74\% & 32.09\%                     & 32.01\% & 30.87\%   \\
100        & 32.52\% & 32.51\%                     & 32.01\% & 30.24\%   \\
200        & 33.12\% & 31.96\%                     & 32.01\% & 30.33\%  
\end{tabular}
\caption{\textbf{Mixed Distribution} Error Rates of TENT, MixNorm (Algorithm \ref{algorithm1}) with and without fixed affine parameters, MixNormBN (Algorithm \ref{algorithm2}) on \textbf{CIFAR-10C} datasets. All methods are tested with 10,000 randomly ordered samples composed of corrupted images from 15 different corruptions types at severity level 5. All methods adopt \textbf{Wide-ResNet-28} as backbone architectures.}
\label{4}
\end{table}

\begin{table}[ht]
\centering
\begin{tabular}{c|cccc}
Batch Size & TENT    & MixNorm w/ Fixed Parameters & MixNorm & MixNormBN \\ \hline
1          & 90.00\% & 15.09\%                     & 15.59\% & 15.62\%   \\
5          & 23.18\% & 14.60\%                     & 15.59\% & 15.61\%   \\
8          & 19.32\% & 14.45\%                     & 15.59\% & 15.46\%   \\
16         & 15.53\% & 14.23\%                     & 15.59\% & 14.26\%   \\
32         & 14.13\% & 14.19\%                     & 15.59\% & 13.53\%   \\
64         & 13.12\% & 14.41\%                     & 15.59\% & 12.78\%   \\
100        & 12.59\% & 13.96\%                     & 15.59\% & 12.32\%   \\
200        & 12.08\% & 13.85\%                     & 15.59\% & 11.95\%  
\end{tabular}
\caption{\textbf{Single Distribution} Error Rates of TENT, MixNorm (Algorithm \ref{algorithm1}) with and without fixed affine parameters, MixNormBN (Algorithm \ref{algorithm2}) on \textbf{CIFAR-10C} datasets. All scores are average error rates from all 15 corruption datasets at severity level 5. All methods adopt \textbf{Wide-ResNet-40} as backbone architectures.}
\label{5}
\end{table}

\begin{table}[ht]
\centering
\begin{tabular}{c|cccc}
Batch Size & TENT    & MixNorm w/ Fixed Parameters & MixNorm & MixNormBN \\ \hline
1          & 89.81\% & 18.80\%                     & 15.99\% & 18.52\%   \\
5          & 27.92\% & 18.80\%                     & 15.99\% & 20.21\%   \\
8          & 24.57\% & 18.80\%                     & 15.99\% & 19.97\%   \\
16         & 21.07\% & 18.80\%                     & 15.99\% & 18.71\%   \\
32         & 18.76\% & 18.80\%                     & 15.99\% & 17.45\%   \\
64         & 17.12\% & 18.80\%                     & 15.99\% & 16.44\%   \\
100        & 17.12\% & 18.80\%                     & 15.99\% & 16.53\%   \\
200        & 16.43\% & 18.80\%                     & 15.99\% & 17.67\%  
\end{tabular}
\caption{\textbf{Mixed Distribution} Error Rates of TENT, MixNorm (Algorithm \ref{algorithm1}) with and without fixed affine parameters, MixNormBN (Algorithm \ref{algorithm2}) on \textbf{CIFAR-10C} datasets. All methods are tested with 10,000 randomly ordered samples composed of corrupted images from 15 different corruptions types at severity level 5. All methods adopt \textbf{Wide-ResNet-40} as backbone architectures.}
\label{6}
\end{table}

\begin{table}[ht]
\centering
\begin{tabular}{c|cccc}
Batch Size & TENT    & MixNorm w/ Fixed Parameters & MixNorm & MixNormBN \\ \hline
1          & 99.87\% & 74.44\%                     & 74.44\% & 74.03\%   \\
5          & 81.08\% & 74.43\%                     & 74.44\% & 74.89\%   \\
8          & 75.74\% & 74.44\%                     & 74.44\% & 72.93\%   \\
16         & 69.85\% & 74.41\%                     & 74.44\% & 69.02\%   \\
32         & 65.51\% & 74.41\%                     & 74.44\% & 65.34\%   \\
64         & 62.74\% & 74.43\%                     & 74.44\% & 62.69\%  
\end{tabular}
\caption{\textbf{Single Distribution} Error Rates of TENT, MixNorm (Algorithm \ref{algorithm1}) with and without fixed affine parameters, MixNormBN (Algorithm \ref{algorithm2}) on \textbf{ImageNet-C} datasets. All scores are average error rates from all 15 corruption datasets at severity level 5. All methods adopt \textbf{ResNet-50} as backbone architectures.}
\label{1}
\end{table}

\begin{table}[ht]
\centering
\begin{tabular}{c|cccc}
Batch Size & TENT    & MixNorm w/ Fixed Parameters & MixNorm & MixNormBN \\ \hline
1          & 99.86\% & 78.88\%                     & 78.89\% & 88.73\%   \\
5          & 90.59\% & 78.87\%                     & 78.89\% & 85.06\%   \\
8          & 88.72\% & 78.87\%                     & 78.89\% & 84.75\%   \\
16         & 87.27\% & 78.85\%                     & 78.89\% & 83.99\%   \\
32         & 86.39\% & 78.87\%                     & 78.89\% & 83.55\%   \\
64         & 86.09\% & 78.86\%                     & 78.89\% & 82.78\%  
\end{tabular}
\caption{\textbf{Mixed Distribution} Error Rates of TENT, MixNorm (Algorithm \ref{algorithm1}) with and without fixed affine parameters, MixNormBN (Algorithm \ref{algorithm2}) on \textbf{ImageNet-C} datasets. All methods are tested with 50,000 randomly ordered samples composed of corrupted images from 15 different corruptions types at severity level 5. All methods adopt \textbf{ResNet-50} as backbone architectures.}
\label{2}
\end{table}

%% file: sections/appendix/hyper.tex
In this section we report the hyper parameters we used for each experiments. As we mentioned in Section 4, we use the ``Gaussian Noise at Severity 5" set to choose the hyper parameters on CIFAR-10C and ImageNet-C experiments. For other dataset, because of the absence of proper validation set, we just pick the hyper-parameters that give the optimal performance. All learning rate stay the same as the learning rated used by other baselines. Finding optimal hyper-parameters remains to be a challenging problem for methods in Test-Time Adaptation. 
\begin{table}[htbp]
\centering
\begin{tabular}{c|ccc}
\multicolumn{1}{l|}{} & Scale $m$ & Moving Speed $\tau$ & Learning Rate \\ \hline
CIFAR10-C Single      & 0.05      & 1.00E-03            & 1.00E-03      \\
CIFAR10-C Mixed       & 0.2       & 1.00E-03            & 1.00E-03      \\
ImageNet-C Single     & 0.01      & 1.00E-03            & 2.50E-04      \\
ImageNet-C Mixed      & 0.05      & 1.00E-06            & 2.50E-04      \\
CIFAR-10              & 0.01      & 1.00E-06            & -             \\
CIFAR-100             & 0.01      & 1.00E-06            & -             \\
STL10                 & 0.01      & 1.00E-06            & -            \\
Stanford-Cars         & 0.01      & 1.00E-07            & -             \\
Food-101              & 0.005      & 1.00E-07            & -             \\
\end{tabular}
\caption{Hyper-Parameters of MixNorm Experiments. In Figure \ref{fig:robustness} and \ref{fig:arch} we reported scores for both MixNorm with and without fixed affine parameters. The MixNorm uses the same hyper-parameter in the both cases.}
\end{table}

\begin{table}[htbp]
\centering
\begin{tabular}{c|ccc}
\multicolumn{1}{l|}{} & Scale $m$ & Max Moving Speed $\tau_{max}$ & Learning Rate \\ \hline
CIFAR10-C Single      & 0.01      & 0.75                          & 1.00E-03      \\
CIFAR10-C Mixed       & 0.2       & 0.75                          & 1.00E-03      \\
ImageNet-C Single     & 0.01      & 1.1                           & 2.50E-04      \\
ImageNet-C Mixed      & 0.05      & 1.1                           & 2.50E-04      \\
MNIST                 & 0.1      & 0.75                          & 0.75      \\
MNIST-M               & 0.01      & 0.75                          & 0.05  \\
USPS                  & 0.01     & 0.5                           & 0.75      \\
\end{tabular}
\caption{Hyper-Parameters of MixNormBN Experiments.}
\end{table}

%% file: sections/appendix/datasets.tex
In Table \ref{dataapp} we report the details of datasets we used for Zero-Shot Classification Experiments.  

\begin{table}[htbp]
\centering
\begin{tabular}{c|ccc}
Datasets      & Classes & Train Size & Test Size \\ \hline
CIFAR-10      & 10      & 50000      & 10000     \\
CIFAR-100     & 100     & 50000      & 10000     \\
Stanford-Cars & 196     & 8144       & 8041      \\
Food-101      & 101     & 75750      & 25250     \\
STL10         & 10      & 1000       & 8000     
\end{tabular}
\caption{Dataset Size and Class Number}
\label{dataapp}
\end{table}